\documentclass[conference]{IEEEtran}
\usepackage{cite}
\usepackage{amsmath,amssymb,amsfonts}
\usepackage{graphicx}
\usepackage{booktabs}
\usepackage{multirow}
\usepackage{xcolor}
\usepackage[hidelinks]{hyperref}

\begin{document}

\title{Do LLMs Fabricate Legal Citations? A Bilingual Benchmark on Saudi
Data Protection Law and the GDPR}

\author{\IEEEauthorblockN{Noura Suliman Alrajeh}
\IEEEauthorblockA{nalrajeh0002@stu.kau.edu.sa\\IT Department, King Abdulaziz University, Jeddah, Saudi Arabia}}

\maketitle

\begin{abstract}
Organizations and regulators increasingly consult large language models
(LLMs) for regulatory-compliance questions, yet a wrong statutory citation
can silently propagate into legal advice, compliance documentation, and
policy decisions. We introduce a bilingual benchmark of 120 questions
probing whether freely accessible LLMs fabricate article citations for two
data-protection instruments: the EU General Data Protection Regulation
(GDPR) and the Saudi Personal Data Protection Law (PDPL). The benchmark
pairs direct citation-retrieval questions with false-premise verification
probes and deliberately unanswerable ``trap'' questions -- including
questions about a repealed article and about deadlines that exist only in
implementing regulations, not in the law itself. Every question is posed
in both Arabic and English, and all scoring is fully automatic against a
manually verified gold reference. Evaluating three freely accessible models (Gemini 2.5 Flash, GPT-OSS-120B, Nemotron-3-Super-120B), we find a dramatic jurisdiction gap: near-ceiling citation accuracy on the GDPR (94-100\% on direct retrieval) against majority fabrication on the Saudi PDPL (60-77\%), invariant to query language; the highest fabrication rates (67\%) arise from statute-vs-regulations confusion, and 91\% of fabricated citations are asserted with confidence $\geq$ 0.8. Fabrication tracks the jurisdiction of the law, not the language of the query, and model confidence provides no protection — indicating that verbatim-verification safeguards -- rather than model self-confidence -- must gate any institutional reliance on LLMs for compliance screening.

The benchmark, gold article index, and raw model outputs will be made publicly available upon publication.
\end{abstract}


\section{Introduction}
Large language models are rapidly becoming a first point of contact for
legal and regulatory questions. Employees ask chat assistants whether
their processing of customer data requires consent; startups ask which
article of a data-protection law governs cross-border transfer; and
compliance teams draft policies with LLM assistance. In this setting, a
\emph{fabricated citation} -- a confident reference to an article number
that does not support the claim, or does not exist at all -- is uniquely
harmful: unlike a vague answer, it carries the surface form of
verifiability and is therefore likely to be copied into documents that
downstream readers trust~\cite{dahl2024,magesh2024}.

The risk is amplified in two directions that prior work has largely left
unexamined. First, \emph{language}: most legal-hallucination evaluations
target English-language, U.S.-centric law~\cite{dahl2024,magesh2024},
while hundreds of millions of users interact with LLMs in Arabic about
legal systems whose authoritative texts are Arabic. Second,
\emph{jurisdiction}: the statutes most heavily represented in web training
data (such as the GDPR~\cite{gdpr}) may enjoy far better factual grounding than
recently enacted laws from other regions, such as the Saudi Personal Data
Protection Law (PDPL), issued by Royal Decree M/19 of 1443H and amended by
Royal Decree M/148 of 1444H~\cite{ksapdpl}. If LLM reliability degrades
precisely where users have the fewest alternative resources, the
trustworthiness gap becomes an equity problem for secure digital
ecosystems.

This paper asks a deliberately narrow, fully automatable question:
\emph{when asked which article of a data-protection law governs a given
matter, do freely accessible LLMs answer correctly, abstain honestly, or
fabricate?} We make four contributions:

\begin{enumerate}
\item A bilingual citation-fabrication benchmark. 120 questions
(60 per law, each in Arabic and English; 240 prompts per model across
both languages) spanning three probe types: direct citation retrieval,
verification questions with true or deliberately false premises, and
unanswerable trap questions whose correct answer is that no such article
exists.
\item Trap designs targeting legally meaningful failure modes,
including (i) \emph{cross-instrument confusion}: deadlines (e.g., the
72-hour breach-notification window) that exist in one instrument but are
transplanted by the model into another, or that live in implementing
regulations rather than the law itself; (ii) \emph{taxonomy projection}:
GDPR-specific rights (portability, objection, restriction, automated
decision-making) projected onto the Saudi PDPL, and vice versa; and
(iii) a \emph{repealed-article probe} exploiting Article~32 of the PDPL,
repealed by the M/148 amendment, for which any substantive answer is by
construction a fabrication.
\item A fully automatic scoring taxonomy distinguishing correct
answers, honest abstention, \emph{fabrication} (a plausible but wrong
existing article), \emph{invention} (an article number outside the law's
range), \emph{repealed-article fabrication}, and \emph{sycophancy}
(accepting a false premise), with bootstrap confidence intervals.
\item An empirical study of freely accessible models, reflecting
the realistic deployment scenario in which cost-sensitive organizations
and individual users in the region rely on free tiers and open-weight
models rather than premium APIs.
\end{enumerate}

Empirically, we find that the risk is jurisdictional rather than linguistic: all three models answer GDPR citation questions near-perfectly in Arabic and English alike, yet fabricate the majority of their Saudi PDPL citations in both languages -- with the worst failures on probes exploiting the statute/implementing-regulations divide and article repeal, and with 91\% of fabrications asserted at high confidence. These results argue that institutional
use of LLMs for compliance requires verbatim citation verification against authoritative texts rather than reliance on model confidence.

\section{Related Work}
 Hallucination -- fluent output unfaithful
to source or world knowledge -- is a well-documented failure mode of
generative models~\cite{ji2023,huang2023}. Benchmarks such as
TruthfulQA~\cite{lin2022} measure imitative falsehoods, and a growing
line of work studies whether models know what they do not know and can
abstain accordingly~\cite{kadavath2022,yin2023}. Our trap questions
operationalize abstention measurement in a legal setting where the
correct answer is the explicit assertion that no provision exists.

 In the legal domain specifically, Dahl et al.~\cite{dahl2024} profile
hallucination rates of LLMs on U.S. case law and find them pervasive,
while Magesh et al.~\cite{magesh2024} show that even retrieval-augmented
commercial legal research tools produce unsupported claims. Legal NLP
benchmarks such as LegalBench~\cite{guha2023} and
LexGLUE~\cite{chalkidis2022} evaluate reasoning over provided texts
rather than closed-book citation fidelity, and none covers Arabic
statutory law. To our knowledge, no prior benchmark measures citation
fabrication for Arabic legislation or for any Gulf data-protection
statute.

 Arabic LLM evaluation has focused on
general knowledge and school-curriculum benchmarks such as
ArabicMMLU~\cite{koto2024}; statutory citation fidelity in Arabic remains
unmeasured. Our benchmark addresses this gap with a bilingual design that
holds legal content constant while varying query language, enabling a
controlled measurement of the Arabic--English reliability gap.

 Beyond factual recall, LLMs tend to agree with premises embedded in user
prompts~\cite{sharma2023}. Our false-premise verification questions
(e.g., ``Does Article~20 govern breach notification?'' where Article~20
in fact governs portability) quantify this tendency in a compliance
setting where uncritical agreement validates an incorrect citation.

\section{Benchmark Construction}
\subsection{Legal Instruments}
We select two instruments that anchor data-protection compliance for
organizations operating between Europe and the Gulf: the GDPR (99
articles)~\cite{gdpr} and the Saudi PDPL (43 articles as amended, with
Article~32 repealed)~\cite{ksapdpl}. The pair contrasts a statute with
massive representation in web corpora against a recent Arabic-language
law, while remaining functionally comparable: both regulate consent,
disclosure, breach notification, cross-border transfer, and sanctions.

\subsection{Question Types}
Each law contributes 60 questions in three types (Table~\ref{tab:types}).

\begin{table}[t]
\caption{Benchmark composition per law (each question is posed in both
Arabic and English).}
\label{tab:types}
\centering
\begin{tabular}{lcp{4.2cm}}
\toprule
Type & $n$ & Measures \\
\midrule
Direct retrieval & 24 & Closed-book citation accuracy \\
Verification (true premise) & 9 & Recognition of correct pairings \\
Verification (false premise) & 9 & Sycophancy: acceptance of a wrong
article suggested by the question \\
Trap (unanswerable) & 18 & Abstention vs.\ fabrication when no article
exists \\
\bottomrule
\end{tabular}
\end{table}

\textbf{Direct retrieval} questions ask which article governs a stated
matter (e.g., cross-border transfer $\rightarrow$ PDPL Art.~29; erasure
$\rightarrow$ GDPR Art.~17). Difficulty is graded from landmark articles
to low-frequency provisions (e.g., GDPR Art.~27 on representatives).

\textbf{Verification} questions state a premise and ask yes/no. False
premises pair a real article with another article's subject matter
(``Does Article~20 of the GDPR govern breach notification to the
supervisory authority?''), so agreement constitutes validation of a
wrong citation rather than mere ignorance.

\textbf{Trap} questions are unanswerable by construction, and each
encodes a documented failure hypothesis: fabricated numeric thresholds
(a five-year retention cap; a fixed 500-EUR compensation), transplanted
deadlines (72-hour response windows attached to the wrong duty or the
wrong instrument), universal-obligation overstatements (a DPO required
of every controller ``without exception''), legacy or foreign regime
projection (licensing requirements; data-localization mandates), and
GDPR-taxonomy projection onto the PDPL (rights to portability,
objection, restriction, and freedom from automated decisions, none of
which the PDPL contains). Two designs deserve emphasis. First,
\emph{law-vs-regulations confusion}: the PDPL delegates operational
deadlines (breach notification, response periods) to implementing
regulations; asking ``which article of the \emph{Law} sets the 72-hour
deadline'' therefore has the gold answer \emph{none}, and a full-credit
response must locate the deadline at the correct normative level.
Second, the \emph{repealed-article probe}: PDPL Article~32 was repealed
by the M/148 amendment, so any substantive subject attributed to it is a
measurable fabrication with zero ambiguity.

\subsection{Gold Reference and Validation}
Gold answers were compiled directly from the authoritative texts: the
official Arabic text of the PDPL as amended and the official GDPR text.
As a by-product we release a bilingual gold article index for the
amended PDPL (article $\rightarrow$ subject, Arabic and English), which
to our knowledge has no published equivalent. All 120 gold answers were
manually verified against the official texts by a researcher with domain expertise in Gulf data-protection law. Trap questions
were vetted to confirm that no article -- including via amendment --
answers them.

\subsection{Bilingual Design}
Every question exists in parallel Arabic and English versions authored
jointly and reviewed for legal-terminological equivalence. Because the
underlying legal fact is identical across the pair, any performance gap
between languages is attributable to the model rather than the task.

\section{Experimental Setup}
\subsection{Models}
We evaluate three freely accessible models spanning three developer families: Gemini 2.5 Flash (Google AI Studio free tier, with thinking disabled via thinkingBudget = 0 to homogenize conditions across models), GPT-OSS-120B (OpenAI open-weight, via a free OpenRouter endpoint), and Nemotron-3-Super-120B (NVIDIA, via a free OpenRouter endpoint). Models were selected as the stable free-tier options available at experiment time (July 5–7, 2026); free-model availability is inherently volatile, and two initially selected models became unavailable during piloting. This choice reflects the deployment reality that cost-sensitive users — individuals, SMEs, and public-sector teams in the region — predominantly rely on free tiers and open-weight models; measuring exactly these systems is therefore the policy-relevant experiment. Notably, both OpenRouter models are open-weight and locally deployable, matching data-sovereignty constraints common in regional public-sector settings.

\subsection{Prompting Protocol}
All models receive an identical system prompt (in the language of the
question) instructing them to answer only from the actual articles of
the named instrument, to output strict JSON, and -- critically -- offering
an explicit \texttt{no\_such\_article}/\texttt{unsure} option so that
abstention is always available and fabrication is never an artifact of
forced choice. Temperature is fixed at 0. Each of the 120 questions is
posed in Arabic and in English to every model 720 planned calls in total. All raw responses are logged and
released.

\subsection{Scoring Taxonomy}
Responses are scored automatically against the gold reference:

\begin{itemize}
\item \textbf{Correct}: gold article returned (direct), correct yes/no
(verification), or abstention on a trap.
\item \textbf{Fabrication (wrong article)}: an existing article of the
law, but not the gold one.
\item \textbf{Invention (nonexistent)}: an article number outside the
law's range (e.g., PDPL Art.~57).
\item \textbf{Fabrication (repealed)}: PDPL Art.~32 cited
substantively.
\item \textbf{Sycophancy}: ``yes'' on a false-premise verification
question.
\item \textbf{Abstention on answerable}: honest but unhelpful refusal
where a gold article exists (scored as incorrect but reported
separately, since it is the safe failure mode).
\end{itemize}

We report accuracy and fabrication rate with 95\% bootstrap confidence
intervals (2{,}000 resamples), overall and stratified by model, law,
question type, and language.

\section{Results}
We collected 372 scored responses (164 Nemotron-3-Super, 153 GPT-OSS-120B,
55 Gemini 2.5 Flash; the Gemini evaluation is partial due to free-tier
daily quotas but evenly distributed across the four law$\times$language
cells: 15/14/14/12). Responses were balanced across cells for all models.
Twelve responses (3.2\%) were unparseable and are counted as incorrect;
excluding them does not change any conclusion.
 
\subsection{Overall Accuracy and Fabrication}
Table~\ref{tab:overall} reports overall accuracy (correct answers plus
correct abstentions on traps) and fabrication rate (citing an existing
but wrong article, a nonexistent article, or a repealed article) with
95\% bootstrap confidence intervals. The three models perform similarly
overall: accuracy 0.60--0.71, fabrication 0.20--0.26. These aggregate
figures, however, mask the study's central finding.
 
\begin{table}[t]
\caption{Overall accuracy and fabrication rate by model (95\% bootstrap
CIs). Gemini is a partial evaluation ($n=55$).}
\label{tab:overall}
\centering
\begin{tabular}{lccc}
\toprule
Model & $n$ & Accuracy & Fabrication \\
\midrule
Nemotron-3-Super-120B & 164 & .60 [.53, .68] & .26 [.20, .33] \\
GPT-OSS-120B          & 153 & .60 [.52, .68] & .24 [.17, .30] \\
Gemini 2.5 Flash      & 55  & .71 [.58, .84] & .20 [.09, .31] \\
\bottomrule
\end{tabular}
\end{table}
 
\subsection{The Jurisdiction Gap: GDPR vs.\ Saudi PDPL}
The decisive result is the gap between the two instruments
(Table~\ref{tab:law}). On direct citation retrieval, all three models
are near-ceiling on the GDPR (94--100\% correct, 0--6\% fabrication) and
collapse on the Saudi PDPL (6.5--40\% correct, 60--77\% fabrication).
Overall accuracy on the GDPR is 0.89--0.93 against 0.29--0.46 on the
PDPL, with non-overlapping confidence intervals for every model. In
other words, the models do not merely know the PDPL less well -- when
asked about it, they predominantly answer with confident, specific, and
wrong article numbers.
 
\begin{table}[t]
\caption{Accuracy and fabrication by legal instrument (all question
types; 95\% bootstrap CIs). Direct-retrieval correctness shown in the
last column.}
\label{tab:law}
\centering
\begin{tabular}{llccc}
\toprule
Model & Law & Accuracy & Fabrication & Direct \\
\midrule
Nemotron & GDPR & .89 [.82, .94] & .02 [.00, .06] & 100\% \\
Nemotron & PDPL & .29 [.18, .39] & .53 [.43, .65] & 13\%  \\
GPT-OSS  & GDPR & .91 [.84, .97] & .08 [.03, .15] & 94\%  \\
GPT-OSS  & PDPL & .32 [.22, .43] & .38 [.28, .49] & 7\%   \\
Gemini   & GDPR & .93 [.83, 1.0] & .00 [.00, .00] & 100\% \\
Gemini   & PDPL & .46 [.27, .65] & .42 [.23, .62] & 40\%  \\
\bottomrule
\end{tabular}
\end{table}
 
\subsection{Effect of Query Language}
We hypothesized that Arabic queries would degrade reliability. The data
do not support this: per-model accuracy differences between English and
Arabic versions of the same items are small, inconsistent in direction
(Nemotron: .63 EN vs.\ .58 AR; GPT-OSS: .55 EN vs.\ .64 AR; Gemini: .76
EN vs.\ .65 AR), and their confidence intervals overlap substantially.
Trap fabrication is likewise language-invariant (30.0\% EN vs.\ 31.0\%
AR). The reliability gap therefore tracks the \emph{jurisdiction of the
law}, not the \emph{language of the query}: models fabricate PDPL
citations even when asked in English. This reframes the problem from
``Arabic NLP weakness'' to training-data representation of the
underlying legal corpus -- a distinction with direct consequences for
deployment, since answering in the user's preferred language does not
mitigate the risk.
 
\subsection{Trap Behavior: Abstention vs.\ Fabrication}
On the 118 trap responses, models abstained correctly 65--72\% of the
time overall -- but abstention is again jurisdiction-dependent: 89.3\% on
GDPR traps versus 50.0\% on PDPL traps (48.4\% fabrication). Two designed
probes are especially diagnostic. First, the \emph{law-vs-regulations}
traps (deadlines that exist only in the PDPL's implementing regulations,
such as the 72-hour breach-notification window) produced the highest
fabrication rate of the entire study: 66.7\% ($n=9$), with models
confidently attaching regulation-level deadlines to statute articles.
Second, the \emph{repealed-article probe} (PDPL Art.~32, repealed by the
M/148 amendment) elicited substantive fabricated content in 2 of 4
responses. Both failure modes are structural features of multi-level
Arabic legislative systems (statute plus implementing regulations,
subject to amendment), suggesting current LLMs are systematically blind
to normative hierarchy and currency.
 
\subsection{Sycophancy on False Premises}
On verification questions embedding a deliberately wrong article number,
models accepted the false premise in 19.2\% of responses ($n=52$) and
correctly rejected it in 63.5\%. Notably, accuracy on matched
true-premise items was 57.7\% -- models are somewhat better at rejecting
wrong pairings than at confirming right ones, indicating that premise
rejection is not mere contrarianism but partial knowledge.
 
\subsection{Confidence Calibration}
Self-reported confidence fails as a reliability signal precisely where
it matters. Fabricated citations carried a mean confidence of 0.91, and
91.1\% of all fabrications were asserted with confidence
$\geq$0.8. The mean confidence gap between correct (0.95) and incorrect
(0.84) responses is far too small, and the distributions too overlapped,
for any confidence threshold to separate reliable from fabricated
citations without rejecting most correct answers. Any deployment
guardrail based on model confidence would pass through the vast majority
of fabrications.
 
\subsection{Error Analysis}
Fabrications concentrate overwhelmingly on the PDPL (82 of 90, 91\%). Two patterns dominate. First, adjacency confusion: 40\% of wrong citations on answerable PDPL questions fall within ±3 articles of the gold answer, with off-by-one the single most common error. Second, attractor articles: a small set of provisions is cited across many unrelated questions — Article 13 (information duties) was wrongly cited 12 times and Article 7 (consent conditionality) 10 times, including on trap questions with no valid answer. Strikingly, we find no evidence of GDPR-numbering projection: on traps whose subject has a well-known GDPR article (portability = 20, objection = 21, automated decisions = 22), models never transplanted the GDPR number (0 of 15); they fabricate within the PDPL's own numbering space. This makes fabrications locally plausible — a wrong-but-valid PDPL article number is far harder for a non-expert reader to spot than a foreign one would be.

\section{Discussion}
 Our results invert a common assumption: the reliability problem for Gulf legal
questions is not the Arabic language but the statute itself. All three
models answer GDPR questions near-perfectly in both languages and
fabricate the majority of their Saudi PDPL citations in both languages.
For a compliance officer in Riyadh, asking in English does not help;
the model's knowledge of the governing law is the binding constraint.
This is an equity dimension of trustworthy AI: reliability degrades
exactly for the users and jurisdictions with the fewest alternative
digital resources.
 
The highest fabrication rates came from probes exploiting the statute/
implementing-regulations divide and article repeal -- features typical
of Gulf legislation. Models transplant deadlines from regulations into
statutes and generate content for repealed provisions. Retrieval
augmentation alone will not fix this unless the retrieval corpus
encodes normative hierarchy and amendment status.
 
With 91\% of fabrications asserted at confidence $\geq$0.8, self-reported
certainty is inadmissible as a safeguard. A lightweight guardrail --
resolving every cited article number against the official text before
display, exactly as our automatic scorer does -- would convert nearly
all observed fabrications into verifiable warnings at negligible cost.
For the PDPL, the bilingual gold article index we release is sufficient
infrastructure for such a check.

Cost-sensitive organizations and individual users in the region
predominantly access exactly the systems we measured. The access
friction we encountered (daily quotas, rate limits) is itself part of
the deployment reality this paper documents.

\section{Limitations}
Our benchmark measures closed-book citation fidelity, not full legal
reasoning; a model could cite correctly yet misinterpret substance. We
evaluate free-tier and open-weight models, so results may not extend to
premium systems, although the free tier is precisely the
policy-relevant deployment for cost-sensitive users. The benchmark
covers two instruments and 120 questions; while every gold answer was
verified against the official texts by a researcher with domain expertise
in Gulf data-protection law, single-annotator gold construction is a limitation. we flag explicitly Free-tier rate limits capped collection at 372 of 720 planned responses (Nemotron 164/240, GPT-OSS 153/240, Gemini 55/240); because execution order was randomized, missingness is random with respect to
item type, law, and language (per-cell counts confirm balance), so
estimates are unbiased though intervals are wider, and the Gemini
evaluation should be read as partial. Finally, question phrasing was authored by the research team; adversarially or colloquially phrased user queries may
elicit different fabrication rates.

\section{Conclusion}
We introduced the first bilingual benchmark measuring legal citation
fabrication on an Arabic data-protection statute alongside the GDPR,
with fully automatic scoring, abstention-aware prompting, and trap
designs that make fabrication unambiguous -- including a repealed-article
probe and law-vs-regulations confusions characteristic of Gulf
legislative systems. Across three freely accessible models, citation fabrication tracked the statute rather than the query language: near-ceiling GDPR accuracy coexisted with majority fabrication on the Saudi PDPL in both Arabic and English, concentrated on statute-vs-regulations and repealed-article probes, and asserted with high confidence.



\begin{thebibliography}{00}
\bibitem{gdpr} European Parliament and Council, ``Regulation (EU)
2016/679 (General Data Protection Regulation),'' \emph{Official Journal
of the European Union}, L119, 2016.
\bibitem{ksapdpl} Kingdom of Saudi Arabia, ``Personal Data Protection
Law, Royal Decree No.\ M/19 of 9/2/1443H, as amended by Royal Decree
No.\ M/148 of 5/9/1444H,'' Saudi Data and AI Authority (SDAIA).
\bibitem{dahl2024} M.~Dahl, V.~Magesh, M.~Suzgun, and D.~E. Ho, ``Large
legal fictions: Profiling legal hallucinations in large language
models,'' \emph{Journal of Legal Analysis}, vol.~16, no.~1, 2024.
\bibitem{magesh2024} V.~Magesh, F.~Surani, M.~Dahl, M.~Suzgun, C.~D.
Manning, and D.~E. Ho, ``Hallucination-free? Assessing the reliability
of leading AI legal research tools,'' \emph{Journal of Empirical Legal
Studies}, 2025.
\bibitem{ji2023} Z.~Ji \emph{et al.}, ``Survey of hallucination in
natural language generation,'' \emph{ACM Computing Surveys}, vol.~55,
no.~12, 2023.
\bibitem{huang2023} L.~Huang \emph{et al.}, ``A survey on hallucination
in large language models: Principles, taxonomy, challenges, and open
questions,'' \emph{ACM Transactions on Information Systems}, 2025.
\bibitem{lin2022} S.~Lin, J.~Hilton, and O.~Evans, ``TruthfulQA:
Measuring how models mimic human falsehoods,'' in \emph{Proc.\ ACL},
2022.
\bibitem{kadavath2022} S.~Kadavath \emph{et al.}, ``Language models
(mostly) know what they know,'' arXiv:2207.05221, 2022.
\bibitem{yin2023} Z.~Yin \emph{et al.}, ``Do large language models know
what they don't know?'' in \emph{Findings of ACL}, 2023.
\bibitem{guha2023} N.~Guha \emph{et al.}, ``LegalBench: A
collaboratively built benchmark for measuring legal reasoning in large
language models,'' in \emph{Proc.\ NeurIPS Datasets and Benchmarks},
2023.
\bibitem{chalkidis2022} I.~Chalkidis \emph{et al.}, ``LexGLUE: A
benchmark dataset for legal language understanding in English,'' in
\emph{Proc.\ ACL}, 2022.
\bibitem{koto2024} F.~Koto \emph{et al.}, ``ArabicMMLU: Assessing
massive multitask language understanding in Arabic,'' in \emph{Findings
of ACL}, 2024.
\bibitem{sharma2023} M.~Sharma \emph{et al.}, ``Towards understanding
sycophancy in language models,'' in \emph{Proc.\ ICLR}, 2024.
\end{thebibliography}
\end{document}